\documentclass{article}
\usepackage{spconf}
\usepackage[utf8]{inputenc} 
\usepackage[T1]{fontenc}    
\usepackage[pagebackref,breaklinks,colorlinks]{hyperref}
\usepackage{url}            
\usepackage{booktabs}       
\usepackage{amsfonts}       
\usepackage{nicefrac}       
\usepackage{microtype}      
\usepackage[table]{xcolor}         
\usepackage{amsmath,amssymb,mathrsfs}
\usepackage{mathtools,wallpaper}
\usepackage{algorithm}
\usepackage{algpseudocode}
\usepackage{bm}
\usepackage{amsthm}
\usepackage{graphicx}
\usepackage{enumitem}
\usepackage{multirow}
\usepackage{tabularx}
\usepackage{makecell}
\usepackage{caption}
\usepackage{float}
\usepackage{arydshln}
\usepackage{diagbox}
\usepackage{subcaption}
\title{Training-Adaptive Convolutional Sparse Coding via Information Bottleneck for Robust Visual Representation}
\ninept
\name{Meng'en Qin\textsuperscript{1}, Yinchen Liu\textsuperscript{2}, Mingxuan Cui\textsuperscript{3}, Youlu Xing\textsuperscript{1,$\ast$}\thanks{$\ast$ Corresponding author. Email: \texttt{mengenching@gmail.com}.}}
\address{
\textsuperscript{1}Faculty of Computer Science and Artificial Intelligence,\\
Shenzhen University of Advanced Technology, Shenzhen, China\\
\textsuperscript{2}School of Mathematical Sciences, \\
University of Electronic Science and Technology of China, Chengdu, China\\
\textsuperscript{3}School of Mathematics, Shandong University, Jinan, China
}

\begin{document}
\topmargin=0mm
\maketitle
\begin{abstract}
Visual signals require compact yet sufficient representations for robust downstream prediction. Convolutional sparse coding (CSC) provides an explicit mechanism for suppressing redundant components while preserving signal content, but its sparsity coefficient is often manually selected during training. We propose a training-adaptive convolutional sparse coding framework for robust visual signal representation. Specifically, we unfold the CSC optimization with the Fast Iterative Shrinkage-Thresholding Algorithm (FISTA) and treat the sparsity coefficient as a differentiable variable jointly learned with the network parameters. From the information bottleneck perspective, this coefficient controls the trade-off between information retention and compression: the sparsity term promotes compact representations, while the reconstruction term together with task loss preserves task-relevant signal content. We further introduce a label-free post-training strategy that adjusts the compression strength for corrupted inputs with the main network parameters fixed. Experiments on CIFAR and ImageNet demonstrate competitive clean-data recognition and greatly improved robustness under different input perturbations. 
\end{abstract}
\section{Introduction}
\label{intro}
Robust visual recognition relies on learning representations that are both sufficient for downstream tasks and compact with respect to the input signal \cite{tishby2000information,tishby2015deepIB,learners}. The information bottleneck (IB) principle \cite{tishby2000information,tishby2015deepIB,hu2024surveyIB} provides a theoretical perspective on this objective: given an input $X$ and a downstream variable $Y$, a great output representation $T$ should retain information relevant to $Y$ while discarding information in $X$ that is irrelevant to $Y$ as much as possible. The corresponding IB objective can be formulated as
\begin{equation}
    \mathcal{F}_{\mathrm{min}}
    [p(t|x)]
    =
    \underbrace{I(T;X)\downarrow}_{\text{ensure compactness}}-\underbrace{\beta I(T;Y)\uparrow}_{\text{ensure sufficiency}},
    \label{eq:ib_objective}
\end{equation}
where $I(\cdot\,;\cdot)$ denotes mutual information, $I(T;X)$ characterizes the amount of the retained input information, $I(T;Y)$ measures its task-relevant information, and $\beta>0$ controls the trade-off between compression and preservation.
From the IB perspective, the forward propagation of deep networks (e.g., ResNet \cite{resnet}, Swin Transformer \cite{swin}, VMamba \cite{vmamba}) can be viewed as a progressive transformation of the input into increasingly task-oriented representations \cite{IB_loss,doesIB}. Ideally, this transformation should remove task-irrelevant information without discarding information necessary for the task. However, modern deep networks typically optimize the final task loss without explicitly controlling the information trade-off of intermediate representations. As a result, the model may suffer from information degradation \cite{yolov9,revcol,deeply_net} when
\begin{equation}
    I(Y;X)
    \geq
    I\!\left(Y;f_{\boldsymbol{\theta_{1}}}^{1}(X)\right)
    \cdots >
    I\!\left(Y;f_{\boldsymbol{\theta_{n}}}^{n}(X)\right)
    \cdots \geq
    I(Y;T),
    \label{eq:information_degradation}
\end{equation}
where $f_{\boldsymbol{\theta_{n}}}^{n}(X)$ denotes the representation obtained after the first $n$ layers; or information redundancy \cite{taskonomy} if
\begin{equation}
    I(X;f_{\boldsymbol{\theta_{1}}}^{1}(X))
    \geq
    \cdots
    I\!\left(X;f_{\boldsymbol{\theta_{n}}}^{n}(X)\right)
    \cdots \geq
    I(X;T) > I(Y;T),
    \label{eq:information_redundancy}
\end{equation}
Such an imbalance between compactness and sufficiency can hinder the model's robustness under input perturbations.

Convolutional sparse coding \cite{convolutional_csc,online_csc,csc_ada_representation} provides an explicit mechanism for controlling the complexity of visual representations while preserving task-relevant information together with the final task loss. CSC models $X=(x)_M \in \mathbb{R}^{M \times H \times W}$ as
\begin{equation}
    X = D * T \doteq (\sum_{i=1}^{C}(d)_{i} * (t)_i)_M,
    \label{eq:csc_decomposition}
\end{equation}
where $*$ is the convolutional operator, $D=(d)_{M\times C} \in \mathbb{R}^{M \times C \times k \times k}$ denotes a convolutional dictionary, and $T=(t)_C \in \mathbb{R}^{C \times H \times W}$ stands for a sparse representation. The sparse code is obtained by balancing reconstruction fidelity and representation complexity:
\begin{equation}
    T^{*}
    =
    \arg\min_{T} 
    \underbrace{\frac{1}{2}\|X-D*T\|_2^2\downarrow}_{\text{information preservation}}
    +
    \underbrace{\lambda\|T\|_1\downarrow}_{\text{representation compression}}.
    \label{eq:csc_objective}
\end{equation}
The reconstruction term encourages preserving the input signal, whereas the $\ell_1$ term suppresses unnecessary parameters. $\lambda > 0$ provides explicit control of the compression strength. Although $\lambda$ in CSC and $\beta$ in the IB objective arise from different optimization formulations, they play analogous roles in adjusting the balance between representation compression and task-relevant information preservation. This property makes CSC with the task loss a natural signal processing mechanism for realizing the compact-sufficient trade-off advocated by the IB principle.

Some studies have tried to integrate CSC into deep networks, achieving competitive visual performance and improved robustness. ML-CSC \cite{convolutional_csc} pioneers the connection between convolutional networks and sparse coding. Res-CSC and MSD-CSC \cite{towards_csc} explain the relation between the multi-layer convolutional sparse coding network and residual network. CSC-CTRL \cite{loop_csc} uses CSC layers to build invertible deep autoencoding models whose performance can compete with tried-and-tested deep generative models. SCN \cite{supervised_csc} trains a deep and end-to-end sparse coding network with a supervised task-driven algorithm via loss backpropagation. SDNet \cite{revisiting_csc} shows that convolutional sparse coding can be integrated with deep networks through differentiable optimization layers, improving model robustness while maintaining computational efficiency.
However, most approaches typically treat $\lambda$ as a pre-selected hyperparameter in training, which may lead to suboptimal compression across different layers and obstruct learning the IB trade-off. This inspires us to model $\lambda$ as a learnable and adaptive variable whose value is determined jointly by the representation and downstream task.

Motivated by this, we propose an information bottleneck-guided training-adaptive convolutional sparse coding framework (\textbf{TA-CSC}). Rather than fixing the sparse coefficient, we make $\lambda$ differentiable within the FISTA \cite{fista} iterations and learn it jointly with the convolutional dictionary and network parameters. The resulting layer provides a clear, layer-wise control of information compression. This design further enables $\lambda$ post-training adaptation using few unlabelled samples under various corruptions. Our contributions are as follows:
\begin{itemize}[leftmargin=*, itemsep=2pt, topsep=0pt, parsep=0pt, partopsep=0pt]    
    \item
    We establish an explicit connection between CSC and the IB principle and provide an interpretable lens for learning compact and sufficient visual representations.
    \item
    We propose an IB-guided and training-adaptive CSC method that makes $\lambda$ an update variable within the unfolded FISTA iterations.
    \item
    We further design an unsupervised post-training loss for $\lambda$ adaptation to improve model robustness under corrupted inputs, while keeping other parameters fixed.
\end{itemize}
\begin{figure*}[ht]
    \centering
    \includegraphics[width=0.95\linewidth]{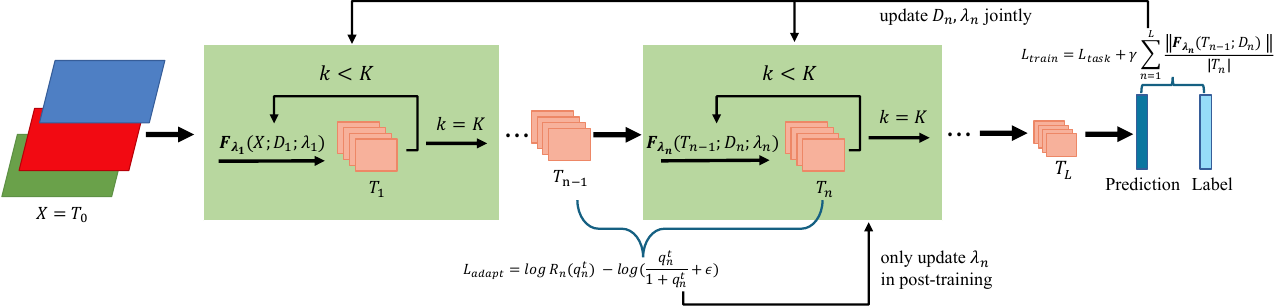}
    \caption{Illustration of the proposed training-adaptive convolutional sparse coding network (\textbf{TA-CSC}).}
    \label{fig:tacsc}
\end{figure*}
\section{Method}
\subsection{Sparse Inference via FISTA}
As shown in Fig. \ref{fig:tacsc}, considering the $k$-th iteration of $n$-th CSC layer, we start the optimization from $T^{(0)}_n=0$, $t^{(1)}_n=1$ and $Y^{(0)}_n=T^{(0)}_n$ \cite{ista}. In Eq. \eqref{eq:csc_objective}, let
\begin{equation}
f(T^{(k)}_n)=\frac{1}{2}\left\|T_{n-1}-D_n*T^{(k)}_n\right\|_{2}^{2},
\end{equation}
whose gradient is Lipschitz continuous with constant $L_n$. In the $k$-th FISTA iteration, we have
\begin{equation}
\begin{aligned}
U^{(k-1)}_n
& =
Y^{(k-1)}_n
-\frac{1}{L_n}\nabla f\!\left(Y^{(k-1)}_n\right), \\
\nabla f\!\left(Y^{(k-1)}_n\right)
& = (D_n)^T * (D_n * Y^{(k-1)}_n - T_{n-1}),
\label{eq:fista_gradient}
\end{aligned}
\end{equation}
followed by
\begin{equation}
T^{(k)}_n
=
\mathcal{S}_{\lambda_n/L_n}\!\left(U^{(k-1)}_n\right),
\label{eq:soft_threshold}
\end{equation}
where the element-wise soft-thresholding operator is defined as
\begin{equation}
\mathcal{S}_{\tau}(u)
=
\operatorname{sign}(u)\max\left(|u|-\tau,0\right).
\label{eq:soft_threshold_def}
\end{equation}
The Nesterov acceleration \cite{fista} is given by
\begin{equation}
\begin{aligned}
Y^{(k)}_n
& =
T^{(k)}_n
+
\frac{t_n^{(k)}-1}{t^{(k+1)}_n}
\left(
T^{(k)}_n-T^{(k-1)}_n
\right), \\
t^{(k+1)}_n
& =
\frac{1+\sqrt{1+4(t_n^{(k)})^2}}{2}.
\label{eq:fista_momentum}
\end{aligned}
\end{equation}
After a fixed number $K$ of iterations, the sparse representation output of $n$-th CSC layer is $T_n=F_{\lambda_n}(T_{n-1};D_n)$.
\subsection{Training-Adaptive Convolutional Sparse Coding}
As discussed in Sec.~\ref{intro}, the sparse coding objective is a favorable choice for the compactness-sufficiency balance advocated by the IB principle. Since $\lambda_n$ directly determines the threshold in \eqref{eq:soft_threshold}, it can be treated as a learnable compression variable. The key is to differentiate the unfolded FISTA iterations with respect to $\lambda_n$.

For the $k$-th iteration, we have 
\begin{equation} 
\frac{\partial T_n^{(k)}}{\partial\lambda_n} = 
\begin{cases} 
-\frac{1}{L_n} \operatorname{sign}(U_n^{(k-1)}) + \frac{\partial U^{(k-1)}_n}{\partial\lambda_n}, & |U_n^{(k-1)}| > \frac{\lambda_n}{L_n}, 
\\[4pt] 
0, & |U_n^{(k-1)}| \leq \frac{\lambda_n}{L_n}. 
\end{cases} 
\label{eq:lambda_t_recursive}
\end{equation}
Because $T_{n-1}$ and $D_n$ are fixed with respect to $\lambda_n$ during the sparse inference of the current layer, differentiating \eqref{eq:fista_gradient} gives
\begin{equation} 
\frac{\partial U^{(k-1)}_n}{\partial\lambda_n} = \left[ I - \frac{1}{L_n}(D_n)^T*D_n \right] * \frac{\partial Y_n^{(k-1)}}{\partial\lambda_n}. 
\label{eq:lambda_u_recursive} 
\end{equation} 
The dependence of the extrapolated variable on $\lambda_n$ is obtained by differentiating \eqref{eq:fista_momentum}. Since $t_n^{(k)}$ is independent of $\lambda_n$, we have 
\begin{equation} 
\begin{aligned} 
\frac{\partial Y_n^{(k-1)}}{\partial\lambda_n} &= \frac{\partial T_n^{(k-1)}}{\partial\lambda_n} + \frac{t_n^{(k-1)}-1}{t_n^{(k)}} \left( \frac{\partial T_n^{(k-1)}}{\partial\lambda_n} - \frac{\partial T_n^{(k-2)}}{\partial\lambda_n} \right). 
\end{aligned} 
\label{eq:lambda_y_recursive} 
\end{equation}
Therefore, with the initialization $\frac{\partial T_n^{(0)}}{\partial\lambda_n} = 
\frac{\partial Y_n^{(0)}}{\partial\lambda_n} = 0$, Eqs.~\eqref{eq:lambda_t_recursive}-\eqref{eq:lambda_y_recursive} provide a complete recursive computation of $\partial T_n/\partial\lambda_n$. For a downstream training loss $\mathcal{L}_{\mathrm{train}}$,
$\frac{\partial\mathcal{L}_{\mathrm{train}}} {\partial\lambda_n} = \frac{\partial\mathcal{L}_{\mathrm{train}}} {\partial T_n} \frac{\partial T_n}{\partial\lambda_n}$.
Consequently, $\lambda_n$ can be optimized jointly with the network parameters by standard backpropagation. Additionally, to enforce non-negativity of $\lambda_n$, we parameterize
their updates as
\begin{equation}
\lambda_n \leftarrow \operatorname{Softplus}(\lambda_n -\eta_{\lambda_n} \cdot \nabla_{\lambda_n}\mathcal{L}_{\mathrm{train}}).
\label{eq:lambda_nonnegative}
\end{equation}

With the hypergradient flow above, we optimize \(\lambda_n\) using the following IB-guided training objective: 
\begin{equation}
\mathcal{L}_{\mathrm{train}}
=
\underbrace{\mathcal{L}_{\mathrm{task}}\downarrow}_{\text{encourage sufficiency}}
+
\underbrace{\gamma \sum_{n=1}^{L}\frac{\|F_{\lambda_n}(T_{n-1};D_n)\|_1}{|T_n|}\downarrow}_{\text{encourage compactness}},
\label{eq:training_loss}
\end{equation}
\begin{equation}
\frac{\partial \mathcal{L}_{\mathrm{train}}}
{\partial \lambda_n}
=
\underbrace{\left\langle
\frac{\partial \mathcal{L}_{\mathrm{task}}}
{\partial T_L},
\frac{\partial T_L}
{\partial \lambda_n}
\right\rangle}_{\text{task gradient}}
+
\underbrace{\gamma
\sum_{m=n}^{L}
\frac{1}{|T_m|}
\left\langle
\operatorname{sign}(T_m),
\frac{\partial T_m}
{\partial \lambda_n}
\right\rangle}_{\text{compression gradient}},
\label{eq:lambda_train_gradient}
\end{equation}
where $\langle , \rangle$ is the inner product, $|T_n|$ denotes the number of elements in \(T_n\), $\mathcal{L}_{\mathrm{task}}$ is the task loss and $\gamma>0$ controls the compression incentive. The task loss encourages preserving task-relevant information, while the second term encourages larger sparsity and stronger suppression of redundant components. Their competition implements the compression-retention trade-off motivated by the IB principle.
\subsection{Label-free Post-training Adaptation for Corrupted Data}
After source-domain training, the network parameters and the sparsity coefficients are denoted by $(\boldsymbol{\theta},\boldsymbol{\lambda})$. When the input distribution is corrupted or shifted, the learned $\boldsymbol{\lambda}$ from clean data may no longer provide an appropriate compression between redundancy removal and task-relevant preservation. We therefore adapt the compression coefficients using a small set of unlabeled corrupted or shifted samples while keeping $\boldsymbol{\theta}$ fixed. We define the relative reconstruction error as a label-free fidelity measure:
\begin{equation}
\mathcal{R}_n
=
\frac{
\left\|\hat{T_{n-1}^t}-D_n*\hat{T_{n}^{t}}\right\|_{2}^{2}
}{
\left\|T_{n-1}-D_n*T_{n}\right\|_{2}^{2} + \epsilon
},
\hat{T_n^t}=F_{\lambda_n^t}(\hat{T_{n-1}^t};D_n),
\label{eq:relative_reconstruction}
\end{equation}
where $\epsilon>0$ is a small constant for numerical stability, $t$ means $t$-th $\lambda_n$ update. We then optimize $\boldsymbol{\lambda}$ using the following label-free loss:
\begin{equation}
\mathcal{L}_{\mathrm{adapt}}
=
\sum_{n=1}^{L}
\left[
\underbrace{\log \mathcal{R}_n(q_n^t)\downarrow}_{\text{reconstruction distortion}}
-
\underbrace{\log\left(
\frac{q_n^t}{1+q_n^t}+\epsilon
\right)\uparrow}_{\text{compression benefit}}
\right],
\label{eq:adapt_loss}
\end{equation}
where $q_n^t=\frac{\lambda_n^t}{\lambda_n^0}>0$. Minimizing \eqref{eq:adapt_loss} encourages larger $\boldsymbol{\lambda}$ to suppress redundant components, while the relative reconstruction term penalizes excessive compression that would distort the observed signal. In contrast to the source-domain training objective \eqref{eq:training_loss}, \eqref{eq:adapt_loss} depends only on the observed corrupted signal and the sparse reconstruction, enabling unsupervised post-training adaptation.
During adaptation, the main network parameters are frozen, and only the coefficients $\boldsymbol{\lambda}$ are updated:
$
\lambda_n^{t+1}
=
\operatorname{Softplus}(\lambda^{t}_n
-
\eta_{\lambda_n^t} \cdot
\nabla_{\lambda_n^t}
\mathcal{L}_{\mathrm{adapt}}).
\label{eq:lambda_adaptation}
$
When batch normalization \cite{batch_norm} is employed, its statistics can be updated using the corrupted batches to adapt to the distribution shift.
\section{Experiments}
We evaluate the proposed TA-CSC on the ImageNet-1K \cite{imagenet}, CIFAR-10 and CIFAR-100 \cite{cifar} datasets. We use ResNet-18 as the baseline backbone and construct two variants: TA-CSC-18 and TA-CSC-18$_{all}$, which replace the first and all convolutional layers with TA-CSC layers, respectively. Two FISTA iterations are unrolled to perform the forward pass of each CSC layer, and $\gamma$ is set to 0.001 in Eq. \eqref{eq:training_loss} across all experiments.
For post-training adaptation, a small unlabeled subset (100 by default) of corrupted samples is used to update the compression coefficient $\boldsymbol{\lambda}$ while keeping $\boldsymbol{\theta}$ frozen. To train models, we used a single NVIDIA RTX 2080Ti with batch size 128 for CIFAR-10/100, and 4 NVIDIA RTX 3090 GPUs with batch size 512 for ImageNet. We compare against ResNet-18 \cite{resnet} and other CSC methods under the same training protocol.
\subsection{Classification Performance on Clean Data}
\begin{figure}[ht]
    \centering
    \includegraphics[width=0.42\linewidth]{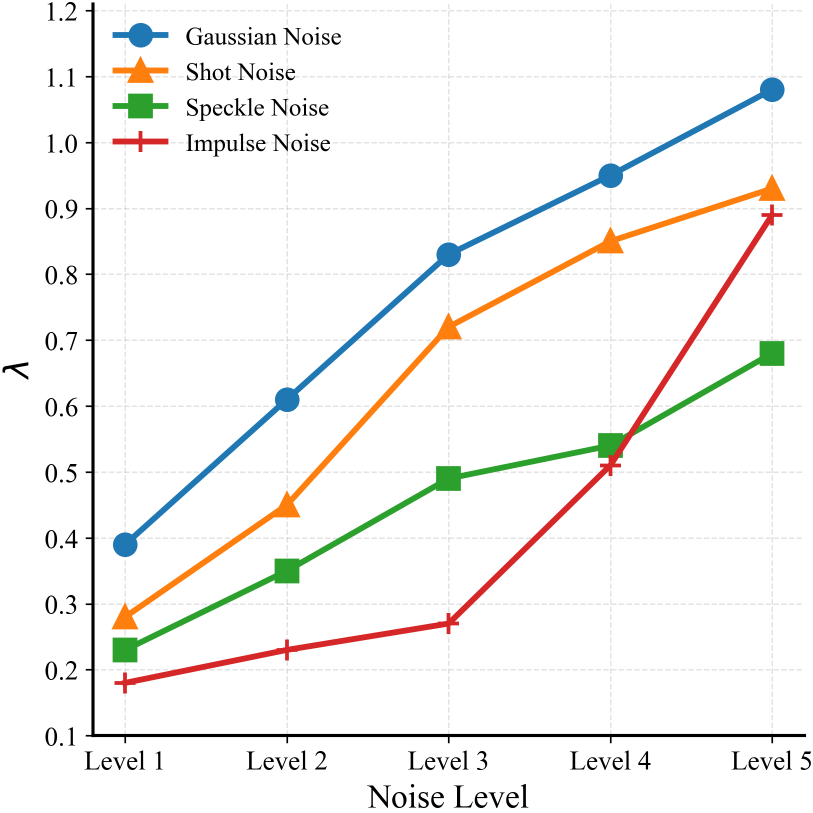}
    \caption{The adapted $\boldsymbol{\lambda}$ value under different noise types and severities.}
    \label{fig:lambda_noise}
\end{figure}
\begin{figure*}[ht]
    \centering
    \includegraphics[width=1.0\linewidth]{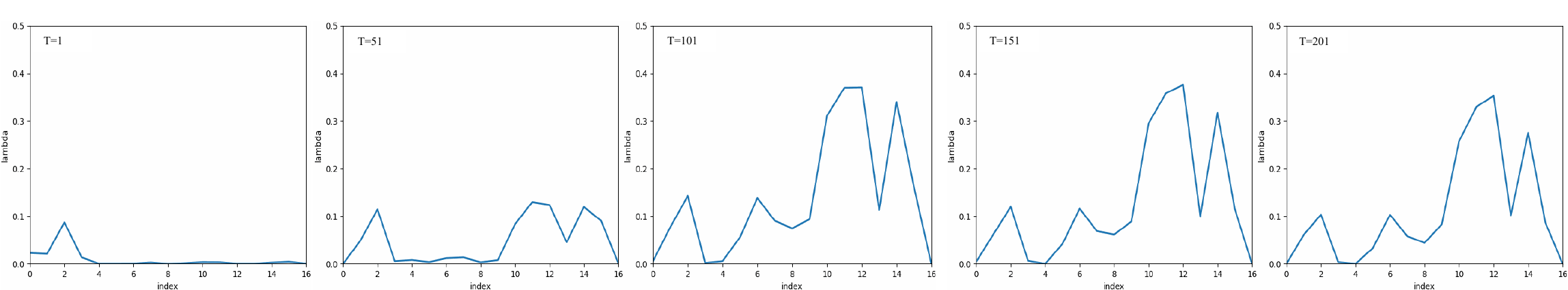}
    \caption{$\boldsymbol{\lambda}$ dynamic behavior when TA-CSC-18$_{all}$ is trained on CIFAR-10. We visualize $\lambda$ values of all TA-CSC layers at the $T$-th iteration.}
    \label{fig:lambda_training}
\end{figure*}
\begin{table}[ht]
    \centering
    \caption{Performance of different methods on clean test data, including CIFAR-10, CIFAR-100, and ImageNet datasets.}
    \label{tab:clean}
    \footnotesize
    \resizebox{0.45\textwidth}{!}{
    \setlength{\tabcolsep}{2pt}  
    \setlength{\extrarowheight}{0pt}
    \renewcommand{\arraystretch}{1.1} 
		\begin{tabular}{l|c|c|c|c}
				\toprule
				Dataset & Method  & Top-1 Acc & Memory & Training Speed \\
				\midrule
                \multirow{5}{*}{\textit{\makecell[c]{CIFAR-10}}}
                    & ResNet-18 \cite{resnet} & 95.54\%  & 1.0 GB & 1600 n/s \\
				& SCN-18 \cite{supervised_csc} & 95.12\% & 3.5 GB & 158 n/s \\
				& SDNet-18 \cite{revisiting_csc} & 95.20\% & 1.2 GB & 1500 n/s \\
                \cdashline{2-5}
				& TA-CSC-18 (ours) & 96.18\% & 1.4 GB & 1324 n/s \\
				& TA-CSC-18$_{all}$ (ours) & 97.65\% & 3.8 GB & 463 n/s \\
				\midrule
                \multirow{5}{*}{\textit{\makecell[c]{CIFAR-100}}}
                    & ResNet-18 \cite{resnet} & 77.82\%  & 1.0 GB & 1600 n/s \\
				& SCN-18 \cite{supervised_csc} & 78.59\% & 3.5 GB & 158 n/s \\
				& SDNet-18 \cite{revisiting_csc} & 78.31\% & 1.2 GB & 1500 n/s \\
                \cdashline{2-5}
				& TA-CSC-18 (ours) & 79.63\% & 1.4 GB & 1324 n/s \\
				& TA-CSC-18$_{all}$ (ours) & 80.76\% & 3.8 GB & 463 n/s \\
				\midrule
                \multirow{5}{*}{\textit{\makecell[c]{ImageNet-1K}}}
                    & ResNet-18 \cite{resnet} & 68.98\%  & 24.1 GB & 2100 n/s \\
				& SCN \cite{supervised_csc} & 70.42\% & 95.1 GB & 51 n/s \\
				& SDNet-18 \cite{revisiting_csc} & 69.47\% & 37.6 GB & 1800 n/s \\
                \cdashline{2-5}
				& TA-CSC-18 (ours) & 71.12\% & 39.7 GB & 1689 n/s \\
				& TA-CSC-18$_{all}$ (ours) & 72.53\% & 88.6 GB & 153 n/s \\
				\bottomrule
		\end{tabular}}
\end{table}
Table~\ref{tab:clean} shows that the proposed TA-CSC achieves competitive performance on clean data. When all convolutional layers are replaced with TA-CSC layers, our method reaches $97.65\%$, $80.76\%$ and $72.53\%$ on CIFAR-10, CIFAR-100 and ImageNet, outperforming other methods.
\subsection{Robustness Analysis on Corrupted Data}
\begin{table}[ht]
    \centering
    \caption{Test accuracy of different models trained on clean data and evaluated on different noise types from CIFAR-10-C and ImageNet-C \cite{cifar_c}. The results are averaged over 5 severity levels for each type.}
    \label{tab:corrupt}
    \footnotesize
    \resizebox{0.48\textwidth}{!}{
    \setlength{\tabcolsep}{2pt}
    \setlength{\extrarowheight}{0pt}
    \renewcommand{\arraystretch}{1.1}
    \begin{tabular}{l|cccc|ccc}
        \toprule
        \multirow{2}{*}{
            \diagbox{\textbf{Method}}{\textbf{Noise Type}}
        }
        & \multicolumn{4}{c|}{\textbf{CIFAR-10-C}} 
        & \multicolumn{3}{c}{\textbf{ImageNet-C}} \\
        \cmidrule(lr){2-8}
        & \textbf{Gaussian}
        & \textbf{Shot}
        & \textbf{Speckle}
        & \textbf{Impulse}
        & \textbf{Gaussian}
        & \textbf{Shot}
        & \textbf{Impulse} 
        \\
        \midrule
        ResNet-18 \cite{resnet}
        & 44.43\% & 57.88\% & 62.16\% & 51.72\% & 22.73\% & 21.78\% & 17.38\% \\
        SCN \cite{supervised_csc}
        & 50.79\% & 62.97\% & 67.45\% & 54.19\% & - & - & - \\
        SDNet-18 \cite{revisiting_csc}
        & 50.58\% & 63.29\% & 67.11\% & 54.13\% & 24.98\% & 23.97\% & 19.12\% \\
        \cdashline{1-8}
        TA-CSC-18 (ours)
        & 51.62\% & 63.95\% & 68.23\% & 54.26\% & 25.63\% & 24.49\% & 20.76\% \\
        TA-CSC-18$_{all}$ (ours)
        & 53.98\% & 65.12\% & 69.87\% & 55.39\% & 27.22\% & 26.33\% & 22.09\% \\
        \midrule
        \makecell[l]{SDNet-18\\+ BN-stat adaptation \cite{bn_adapt}}
        & 64.91\% & 71.11\% & 71.39\% & 57.45\% & 29.12\% & 27.55\% & 22.04\% \\
        \cdashline{1-8}
        \makecell[l]{SDNet-18\\+ post-training $\boldsymbol{\lambda}$ adaptation}
        & 65.26\% & 71.84\% & 71.67\% & 58.33\% & 29.31\% & 27.89\% & 22.23\% \\
        \cdashline{1-8}
        \makecell[l]{TA-CSC-18 (ours)\\+ post-training $\boldsymbol{\lambda}$ adaptation}
        & 66.79\% & 72.24\% & 71.93\% & 59.22\% & 29.95\% & 28.15\% & 22.67\% \\
        \cdashline{1-8}
        \makecell[l]{TA-CSC-18$_{all}$ (ours)\\+ post-training $\boldsymbol{\lambda}$ adaptation}
        & 68.34\% & 73.94\% & 72.81\% & 60.37\% & 30.96\% & 29.64\% & 23.53\% \\
        \bottomrule
    \end{tabular}}
\end{table}
Table~\ref{tab:corrupt} shows that, without post-training adaptation, TA-CSC-18 already consistently outperforms both ResNet-18 and other CSC baselines. More importantly, label-free post-training adaptation of $\boldsymbol{\lambda}$ further improves the robustness of the proposed models across all corruption types. 
The SDNet-18 with one-layer $\boldsymbol{\lambda}$ adaptation also surpasses the corresponding SDNet-18 with all-layer BN-stat adaptation \cite{bn_adapt}, demonstrating that the proposed post-training compression coefficient adaptation is more effective than directly tuning BN statistics of a fixed model. These results support the view that robustness can be improved by re-estimating the compression strength according to the corrupted data distribution rather than keeping a fixed compression level learned from clean data.
In Fig.~\ref{fig:lambda_noise}, a monotonic trend can be observed across all noise types: $\boldsymbol{\lambda}$ increases as the corruption severity becomes stronger, indicating that the proposed TA-CSC-18 automatically imposes stronger sparsity constraints when the input contains more redundancy. This behavior is consistent with the information bottleneck interpretation, where $\boldsymbol{\lambda}$ serves as a controllable compression variable that increases the suppression of task-irrelevant components as the amount of nuisance information grows.
\begin{table}[ht]
    \centering
    \caption{Ablation on the number of FISTA iterations and number of corrupted samples used in post-training. We report TA-CSC-18 performance on CIFAR-10 \cite{cifar} and CIFAR-10-C \cite{cifar_c}.}
    \label{tab:ablation}
    \footnotesize
    \resizebox{0.48\textwidth}{!}{
    \setlength{\tabcolsep}{2pt}  
    \setlength{\extrarowheight}{0pt}
    \renewcommand{\arraystretch}{1.1} 
    \begin{tabular}{cc|ccccc}
    \toprule
     iteration number & Top-1 Acc & \makecell[c]{sample number} & Gaussian & Shot & Speckle & Impulse \\
    \midrule
    2 & 96.18\% & 50 & 66.14\% & 76.95\% & 71.22\% & 58.75\% \\
    4 & 96.54\% & 100 & 66.79\% & 72.24\% & 71.93\% & 59.22\% \\
    8 & 96.93\% & 500 & 67.58\% & 73.19\% & 72.54\% & 60.67\% \\
    \bottomrule
    \end{tabular}}
\end{table}
\subsection{Ablation Analysis}
Table~\ref{tab:ablation} studies the effects of the number of FISTA iterations and corrupted samples used for post-training adaptation. Increasing FISTA iterations gradually improves the clean Top-1 accuracy, while increasing the adaptation samples from 50 to 500 also promotes robustness across all corruption types. However, the gains are relatively limited compared with the additional computational costs. We therefore use 2 FISTA iterations and 100 corrupted samples as the default setting.
\subsection{\texorpdfstring{$\boldsymbol{\lambda}$}{Lambda} Dynamic Behavior Analysis in Training}
To investigate how the proposed adaptive compression mechanism evolves during training, we visualize the layer-wise dynamics of $\boldsymbol{\lambda}$ in TA-CSC-18$_{all}$, as is shown in Fig.~\ref{fig:lambda_training}. At the early stage of training, the learned $\boldsymbol{\lambda}$ remains relatively small, allowing the network to preserve more information from the input while primarily optimizing the downstream task. As the training accuracy approaches saturation, the compression coefficients increase rapidly and subsequently converge. This behavior suggests a two-stage learning process: the early training is dominated by task fitting, whereas the later stage increasingly favors the removal of redundant representation components. Such a fitting-compression transition is consistent with the information bottleneck interpretation, which emphasizes retaining task-relevant information while progressively suppressing information that is less useful for the downstream task.
The layer-wise distribution of $\boldsymbol{\lambda}$ further reveals a clear depth-dependent compression pattern. The coefficients in earlier layers are generally smaller, whereas larger values are observed in layers closer to the downstream task. This observation is consistent with the IB view that early layers should preserve a broader range of input information, while representations closer to the prediction objective can impose stronger compression once task-relevant information has been extracted. Therefore, the learned $\boldsymbol{\lambda}$ profile provides an explicit, interpretable indicator of how compression is distributed across the network hierarchy.

In addition, the $\boldsymbol{\lambda}$ exhibits four pronounced compression cycles during training, manifested as four major peaks in the profile. This behavior may be associated with the architecture of ResNet, where the feature representation width is expanded at four stages. Each expansion increases the representational capacity and may consequently introduce additional redundant components. The model responds by assigning stronger compression coefficients around these expansion stages, resulting in the four observed peaks. This architecture-dependent pattern further indicates that the adaptive sparsity coefficients are not merely free parameters, but reflect the representation-compression behavior of different stages of the network.
\section{Conclusion}
We presented an information bottleneck-driven training-adaptive convolutional sparse coding for robust visual signal representation. By unfolding FISTA, $\boldsymbol{\lambda}$ becomes a training variable, enabling the network to jointly learn task-related representations and adaptive compression. We further introduced a label-free post-training adaptation strategy that re-estimates the compression strength for corrupted inputs. 
Despite these results, the current study is mainly evaluated on the ResNet architecture and classification task, and the relationship between the learned $\boldsymbol{\lambda}$ and information compression is supported primarily by empirical evidence. 
Future work will investigate more diverse visual tasks and distribution shifts, establish a more rigorous theoretical connection between sparse coding and the information bottleneck objective, and explore more general adaptive compression mechanisms beyond the CSC architecture.
\newpage
\bibliographystyle{IEEEbib}
\bibliography{refs}
\end{document}